\documentclass[a4paper]{article}

\usepackage{INTERSPEECH2015}

\usepackage{amsmath,graphicx,amsfonts,epsfig,epstopdf,datetime,algorithm,algorithmic,multirow,xcolor,subfigure,slashbox}

\ninept
\title{Recognize Foreign Low-Frequency Words with Similar Pairs}
\makeatletter
\def\name#1{\gdef\@name{#1\\}}
\makeatother \name{{\em Xi Ma$^{1}$, Xiaoxi Wang$^{1}$, Dong Wang$^{*1,2}$, Zhiyong Zhang$^{1}$}}

\address{$^1$CSLT, RIIT, Tsinghua University \\
  $^2$TNList, Tsinghua University \\
  {\small \tt \{mx,wxx\}@cslt.riit.tsinghua.edu.cn}\\
  {\small \tt wangdong99@mails.tsinghua.edu.cn}\\
  {\small \tt zhangzy@cslt.riit.tsinghua.edu.cn}
}

\begin{document}
%
\maketitle
\begin{abstract}
Low-frequency words place a major challenge for automatic speech recognition (ASR). The probabilities of these
words, which are often important name entities,
are generally under-estimated by the language model (LM) due to their limited occurrences in the training data. Recently, we proposed
a word-pair approach to deal with the problem, which borrows information of frequent words to enhance
the probabilities of low-frequency words. This paper presents an extension to the word-pair method by involving multiple `predicting words' to
produce better estimation for low-frequency words. We also employ this approach to deal with out-of-language words in the task of
multi-lingual speech recognition.

\end{abstract}
\section{Introduction}
\label{sec:intro}
The language model (LM) is an important module in automatic speech recognition (ASR).
The most well-known language modelling approach is based upon word n-grams, which relies on statistics of n-gram counts to predict the probability of a word given its past n-1 words.
In spite of the wide usage, the n-gram LM possesses an obvious limitation
in estimating probabilities of words that are with low frequencies
and the words that are absent in the training data.
For low-frequency words, the probabilities tend to be under-estimated due to the
lack of occurrences of their n-grams in the training data; for words that are absent in training,
estimating the probabilities is simply impossible. Ironically, these words are often
important entity names that should be emphasized in decoding, which means the
probability under-estimation for them is a serious problem for ASR systems
in practical usage.

A well-known approach to dealing with low-frequency and absent words is various
smoothing techniques such as back-off~\cite{katz1987estimation} and
discount~\cite{kneser1995improved,chen1996empirical}. This approach
allocates a small proportion of probability mass to low-frequency and absent words
so that they are allowed to be recognized. However, the allocated probabilities
for these words are very small, which makes it unlikely to be well recognized
unless the acoustic evidence is fairly strong. Besides, this approach does
not support flexible enhancement for words that are important for a particular domain
or application.

Another famous approach is to train an LM with some structures that can be dynamically changed, e.g., the
class-based LM with classes that are adaptable online~\cite{ward1996class}. These dynamic structures, however,
need to be pre-defined and can not handle words that are not in the structure.
For example, words that are not in the pre-defined classes cannot be handled by
class-based LMs. Additionally,  involving such dynamic structures often
requires to modify the decoder, which is not ideal to our opinion.

Recently, we proposed a similar-pair approach to deal with the problem~\cite{ma15}.
The basic idea is to borrow some information from high-frequency words to enhance
low-frequency words. More specifically, we seek for a
high-frequency word that is similar to the word to enhance, and then re-weight
the probability of the low-frequency word by adding a proportion of the probability
of the high-frequency words to the the probability of the low-frequency words.
This approach has been implement with the LM FST graph~\cite{mohri2002weighted}. Compared
to the traditional class-based LM approach, the new approach is flexible to enhance any
words and does not need to change the decoder. It has been shown that this approach
can provide significant performance gains for low-frequency words and words that are
totally absent in the training data.

This paper is a following work of~\cite{ma15}. We first present an extension that allows multiple
high-frequency words (`indicating words') to be used when enhancing a low-frequency word. This extension
helps to involve multi-source information in the word enhancement, and is particular important
for words with multiple senses. Secondly, the similar-pair approach is applied to deal with
a particular low-frequency words: out-of-language (OOL) words that are from another language
but embedded in utterances of the host language, for example English words appearing in Chinese utterances.
These words are totally new for the host language and no context information can be employed
to estimate the probabilities for them. The similar-pair approach can deal with the situation,
by assuming that words in different languages share the same semantic space and hence similar pairs
can essentially cross languages. The experimental results in Section~\ref{sec:experiment} demonstrated
the capability of this approach in dealing with OOL words.

The remainder of this paper is structured as follows. Section~\ref{sec:related} discusses relevant works, and the similar-pair method is described in section~\ref{sec:pair}. The two new extensions are presented in Section~\ref{sec:method}, which is followed by Section~\ref{sec:experiment} where the experiments are presented. Finally, the entire paper is concluded by Section~\ref{sec:conclusion}.

\section{Related works}
\label{sec:related}

This work is related to dynamic language modeling that adds new words and re-weighting word
probabilities, particulary the approaches that are  based on FSTs. This section reviews
some typical techniques of this approach, and primarily focuses on the class-based LM that
deals with dynamic vocabularies and low-frequency words.

The class-based language modeling~\cite{brown1992class} is an approach that clusters similar words into classes and
the probabilities of words in each class are re-distributed, for instance according to their unigram statistics.
Typically, the class-based LM delivers better representations than the word-based LM for low-frequency words ~\cite{ward1996class}, since the class-based structure factorizes probabilities of low-frequency words into class probabilities and class member
probabilities, and so increases robustness of the probability estimation. Moreover, new words can be easily
added into classes with the class-based LM, leading to a dynamic vocabulary.
Additionally, \cite{georges2013transducer} and~\cite{schalkwyk2003speech} introduced two techniques
to build both the class-based LMs and the class words into FST graphs and embed class FSTs into the
class-based LM FST. This embedding can be done on-the-fly, thus offers a flexible dynamic decoding
that supports instant introduction of new words. Similar approaches have been proposed
in~\cite{dixon2012specialized, naptali2012topic, samuelsson1999class}, where various dynamic
embedding methods are introduced, and the classes are extended to complex grammars.

The work is an extension of the similar-pair method proposed in~\cite{ma15}. In this approach,
the probabilities of low-frequency words are enhanced and new words are supported by adding new
FST transitions, both referring to the transitions of the similar and high-frequency words.
Compared to the other  approaches mentioned above, this method is more flexible, which
supports any words instead of words limited in some pre-defined classes.

The extensions we made in this paper for the work in~\cite{ma15} are two-fold: firstly, the similar-pair
algorithm is extended to allow multiple predicting words, which enables multiple information engaged and thus
better enhancement; second, the similar-pair approach is employed to deal with OOL words, which demonstrated
that similar pairs can be cross-lingual.

\section{Word enhancement by similar-pairs}
\label{sec:pair}
In this section we first give a brief introduction to the FST-based ASR architecture, and then present the similar pair method implemented on FSTs.

\subsection{Finite state transducer}
\label{ssec:wfst}
\begin{figure}[htb]
  \center
  \includegraphics[width=5cm]{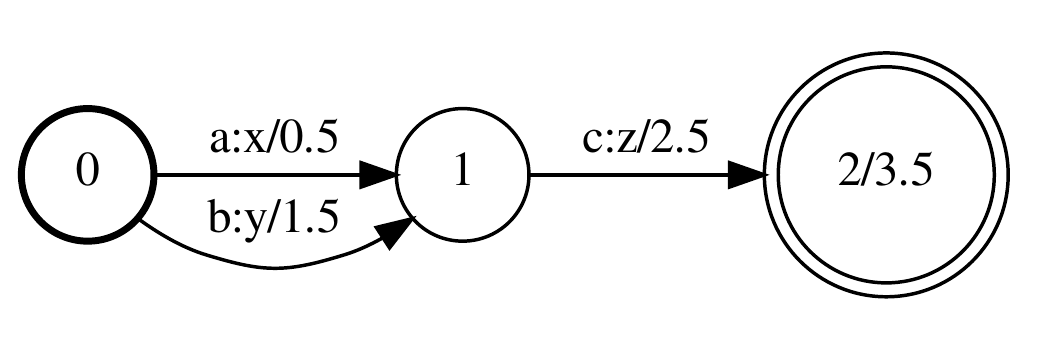}\\
  \caption{An FST example.}\label{fig:wfst}
\end{figure}

A Finite State Transducer (FST) essentially is a Finite State Automaton (FSA) which produces output as well as reading input. It is represented as a graph where every node indicates a state and every arc that links two nodes is assigned an input and an output symbol. Each transition and each terminated state is labeled with a weight. An FST example is depicted in Fig~\ref{fig:wfst}. In this example, the initial state is state $0$, and the final state is state $2$. A weight $3.5$ has been assigned to the final state. Let $(s,t,i:o/w)$ denotes a transition, where $s$ and $t$ are the entry and exiting states respectively, and $i$ is the input symbol and $o$ is the output symbol, and $w$ is the associated transition weight. From the initial state $0$ to state $1$, there are two transitions $($0,1,a:x/0.5$)$ and $($0,1,b:y/1.5$)$. From the state $1$ to the final state $2$, there is only one transition $($1,2,c:z/2.5$)$. An FST can accept a sequence of input symbols and generate a sequence of corresponding output symbols. For instance in Fig.~\ref{fig:wfst}, given an input string \lq ac\rq, the transition $($0,1,a:x/0.5$)$ accepts the first character `a' and generates an output `x' with weight $0.5$, and the transition $($1,2,c:z/2.5$)$ accepts the second character `c' and generates an output `z' with weight $2.5$. The weight of the transition path is computed as the sum of the weights associated to each transition in the path, plus the weight associated with the finally state. In our example, the weight of the transition path that accepts `ac' is $6.5$.

\subsection{FST-based speech recognition}
\label{ssec:hclg}
Most of current large-vocabulary speech recognition systems are based on statistical models, such as hidden Markov models (HMMs), lexicons, decision trees and n-gram LMs. All these models can be converted into the FST models. For an FST, the correlation between the input and output symbols will represent the mapping from a low-level sequence (e.g., phones) to a high-level sequence (e.g., words), and the weights will encode the probability distribution of the mapping. More importantly, FSTs that represent different levels of statistical models can be composed together to form a unified mapping function that associates primary inputs to high-level outputs. The composed FST can be further optimized by standard FST operations, including determinization, minimization and weight pushing. This produces very compact and efficient graphs that represent the knowledge of all the statistical models involved in the composition. In speech recognition, the composition can be used to produce a very efficient graph that maps HMM states to word sequences. The graph building process can be represented as follows:
\begin{equation}\label{graph}
    HCLG = min(det(H \circ C \circ L \circ G))
\end{equation}
where H, C, L and G represent the HMM, the decision tree, the lexicon and the LM (or grammar in grammar-based recognition) respectively, and $\circ$, `$det$' and `$min$' denote the FST operations of composition, determinization and minimization respectively.

\subsection{Low-frequency word enhancement with similar pairs}
\label{ssec:similar}
The similar pairs method is based on the FST architecture. In order to enhance low-frequency words, and for conducting the enhancement on the LM FST, or the $G$ graph, a list of manually defined similar pairs are provided with corresponding frequency information obtained from training data. The low-frequency words are selected to be enhanced and the high-frequency words are chosen to provide the enhancement information. Each similar pair in the list includes one high-frequency word and some low-frequency words.
Given a set of similar words, the low-frequency words are enhanced by looking at the information of the high-frequency word, including its transitions in the $G$ graph and the associated weights. The high-frequency words are preserved since they have been well represented by the n-gram model already.

\section{The Method}
\label{sec:method}
The extension of similar pairs method are introduced in this section. Based upon the similar pairs method, the probability of low-frequency or new words are enhanced by looking at the information of high frequency words. Given a set of words $W=\{x_1, x_2,..., x_m\}$ to be enhanced, for each word $x_i \in W$, a set of words $S_i=\{y_1, y_2,..., y_n\}$ that are similar to $x_i$ is manually selected from the training data.  The similarity can be defined in terms of either syntactic roles or semantic meanings. We assume that, for each $y_j\in S_i$, if there exist an n-gram of $y_j$  in the training corpora, the corresponding n-gram of $x_i$ should also have a relative higher probability of appearance. As the probabilities are represented as the weights in the $G$ graph in FST, according to this assumption, the new weight (probability) of $x_i$ can be updated by the equation (\ref{eq:newweight}).
\begin{equation}
\label{eq:newweight}
    w_{x_i} = w_{y_j} + ln(\frac{f_{x_i}}{f_{x_i}+f_{y_j}}) + \theta
\end{equation}
where $\theta$ is a parameter that tunes the enhancement scale. Note that according to (\ref{eq:newweight}), a larger $f_{x_i}$ leads to a higher $w_{x_i}$, which means that a more frequent word (still low-frequency) is assigned a larger weights after enhancement, and so the rank of the low-frequency words in probabilities is preserved. However, if the word $x_i$ is a new word, the logarithm term can be ignored.
Then the FST can be updated with the new weight.
Let $A(y_j)$ denote the set of all the transitions of the word $y_j$. $A(y_j)$ can be retrieved by searching through the $G$ graph.
For each transition $(s, t, y_j:y_j/w_{y_j}\in A(y_j)$ in the $G$ graph, check if a transition $(s,t,x_i:x_i/w_{x_i})$ exist in $G$ for $x_i$. If it exists, the wight $w_{x_i}$ will be adjusted to a new weight $w'_{x_i}$, otherwise, a parallel arc of transition $(s,t,x_i:x_i/w'_{x_i})$ will be added. The new weight $w'_{x_i}$ can be calculated by the equation (\ref{eq:newweight}).

\begin{figure}[htb]
%
\subfigure[ ]{
  \centering
    \includegraphics[width=4.0cm]{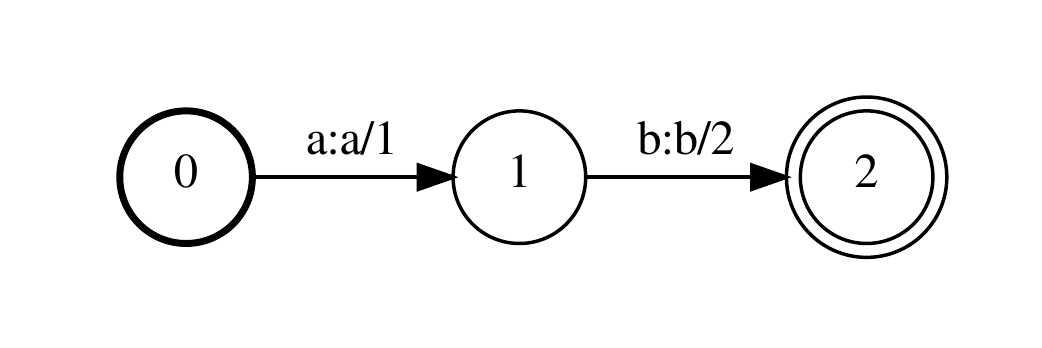}
}
\subfigure[ ]{
  \centering
    \includegraphics[width=4.0cm]{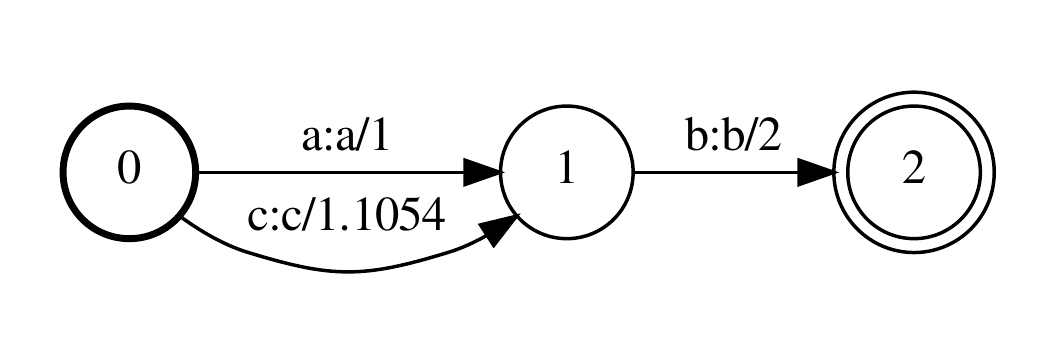}
}
%
\caption{An example of a low-frequency word enhancement based on similar pairs. (a,c) is a similar pair, where `a' is the high-frequency word, and `c' is a low-frequency word. A new transition is added in (b).}
\label{fig:res}
\end{figure}

An example of the enhancement process is illustrated in Fig.~\ref{fig:res}, where Fig.~\ref{fig:res}(a) shows the $G$ graph before the enhancement, and Fig.~\ref{fig:res}(b) shows the $G$ graph after the enhancement. Note that `a' is the high-frequency word, and (a,c) forms a similar pair. A new transition has been added in Fig.~\ref{fig:res}(b) for the low-frequency word $c$.

Comparing with the original similar pairs method, in this extension method each low-frequency or new words $x_i$ will be enhanced by multiple high-frequency words rather than only one. This will increase the search space of the low-frequency word in the $G$ graph. Since the corresponding high-frequency referral words in $S_i$ are selected to be similar to $x_i$, the paths of $x_i$ which are added to $G$ are most likely to be correct in grammar and reasonable in meaning.

In addition, multiple referral words provide more nearly complete information especially in multi-lingual situation, since the one-to-one correspondence are merely existed in few occasions. The usage of a word in one language could have multiple variants in another language. It has been verified by our experiment results in the section \ref{sec:experiment}.

\section{Experiment}
\label{sec:experiment}
The bilingual ASR tasks in the telecom domain is chosen to evaluate the proposed approach. We first introduce the experimental configurations, and then present the performance with the proposed low-frequency English words enhancement based on similar pairs.

\subsection{Database}
\label{ssec:data}

Our ASR task aims to transcribe conversations recorded from online service calls. The domain is the telecom service and the language is in Chinese or English. The acoustic model (AM) is trained on an 1400-hour online speech recording which is manually transcribed from a large call center service provider. The Chinese LM is trained on a corpus including the transcription of the AM training speech and some logs of web-based customer service systems in the domain of telecom service.

22 similar pairs are selected to evaluate the performance of the similar-pair method. Each similar pair contains 1$\sim$5 high-frequency Chinese words and 1$\sim$4 new English words. A `FOREIGN' test set was deliberately designed to test the enhancement with these similar pairs, which consists of $42$ sentences from online speech recording. For each transcription, some new English words that appear in the similar pairs are involved.

Additionally, a `GENERAL' test set that involves $2608$ utterances is selected as a control group to test the generalizability of the proposed method. Each utterance in this set contains words in various frequencies and therefore it can be used to examine if the proposed method impacts general performance of ASR systems at the time of enhancing low-frequency and new words.

\subsection{Acoustic model training}
\label{ssec:am}
The ASR system is based on the state-of-the-art HMM-DNN acoustic modeling approach, which represents the dynamic properties of speech signals using the hidden Markov model (HMM), and represents the state-dependent signal distribution by the deep neural network (DNN) model. The feature used is the 40-dimensional FBank power spectra. A 11-frame splice window is used to concatenate neighboring frames to capture long temporal dependency of speech signals. The linear discriminative analysis (LDA) is applied to reduce the dimension of the concatenated feature to $200$.

The Kaldi toolkit~\cite{Povey_ASRU2011} is used to train the HMM and DNN models. The training process largely follows the WSJ s5 GPU recipe published with Kaldi. Specifically, a pre-DNN system is first constructed based on Gaussian mixture models (GMM), and this system is then used to produce phone alignments of the training data. The alignments are employed to train the DNN-based system.

\subsection{Language model training}
\label{ssec:lm}

The training text is normalized before training. The normalization includes removing unrecognized characters, unifying different encoding schemes and normalizing the spelling form of numbers and letters. Then the training text is segmented into word sequences. A word segmentation tool provided by Google is used in this study. There are totally150,000 words are selected as the LM lexicon, according to the word frequency in the segmented training text. The SRILM toolkit~\footnote{http://www.speech.sri.com/projects/srilm/} is then used to train a 3-gram LM, which is smoothed by Kneser-Ney discounting. The Kaldi toolkit is used to convert n-gram LMs to G graphs, and the openFST toolkit\footnote{http://www.openfst.org} is used to manipulate FSTs.

\subsection{Experiment result and analysis}
\label{ssec:result}

\begin{figure}[htb]
  \includegraphics[width=6cm]{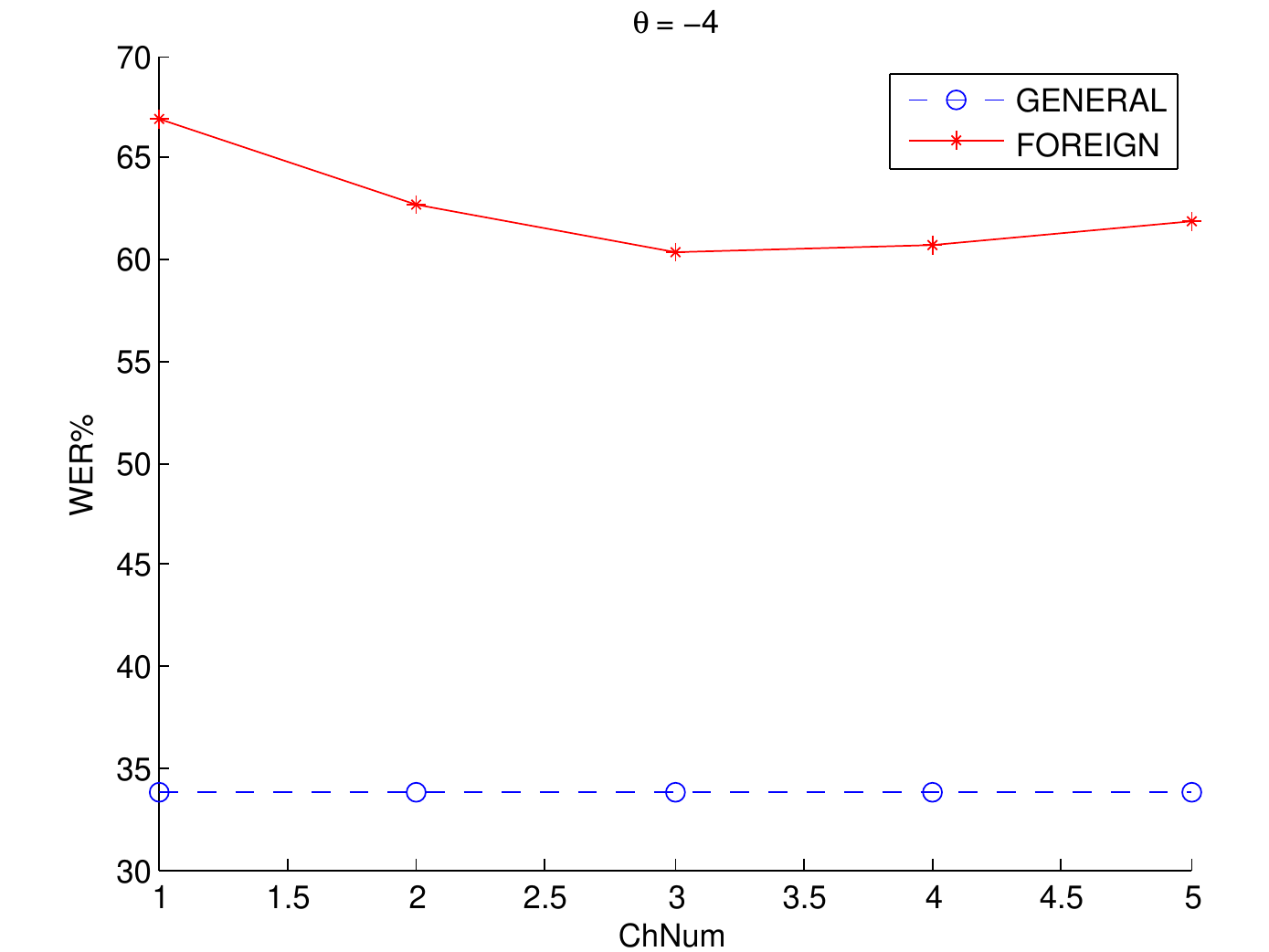}\\
  \caption{WERs on the two test sets with increasing amounts of $ChNum$. The value of $\theta$ is -4}\label{fig:wer}
\end{figure}

The ASR performance in terms of the word error rate (WER) is presented in Table~\ref{tab:wer}, Table~\ref{tab:wer_general} and Table~\ref{tab:wer_foreign}. We report the results on two test sets: `GENERAL' and `FOREIGN', and the results with different values of the enhancement scale $\theta$ and different amounts of high-frequency Chinese words are presented. It can be seen that with the similar-pair-based enhancement, the ASR performance on utterances with new English words is significantly improved. In addition, comparing with the approach of one high-frequency Chinese word, the approach of multiple high-frequency Chinese words has better results. Interestingly, the enhancement on these infrequent words does not cause degradation on the other Chinese words in `GENERAL' test set. Moreover, the performance on the `GENERAL' test set nearly remains unchanged, which indicates that the proposed approach does not impact general performance of ASR systems, and thus is safe to employ. For a more clear presentation, the trends of WERs on the two test sets with different values of $\theta$ and different amounts of high-frequency Chinese words are presented in Fig.~\ref{fig:wer}.

\begin{table}[htb]
\center
\begin{tabular}{l|c|c|c}
  \hline
   & &\multicolumn{2}{c}{WER\%} \\
   \hline
   & $\theta$ & GENERAL & FOREIGN\\
   \hline
  Baseline & - & 33.75 & 77.64 \\
  \hline
  + SP & -4 & 33.76 & 66.95 \\
  \cline{2-4}
     & -2 & 33.76 &  64.39\\
  \cline{2-4}
     & 0 & 33.77 & 62.11 \\
  \cline{2-4}
     & 2 & 33.8 & 62.96 \\
  \cline{2-4}
     & 4 & 33.83 & 69.8 \\
  \hline
\end{tabular}
\footnotesize
\caption{WERs with and without the similar-pair-based enhancement. `SP' stands for enhancement with similar pairs, which use one high-frequency Chinese word. $\theta$ is the enhancement scale in equation(\ref{eq:newweight}).}
\label{tab:wer}
\end{table}

\begin{table}[htb]
\center
\begin{tabular}{c|c|c|c|c|c}
  \hline
    & \multicolumn{5}{c}{WER\%} \\
  \hline
   \backslashbox{$\theta$}{$ChNum$} & 1 & 2 & 3 & 4 & 5 \\
  \hline
   -4 & 33.76 & 33.77 & 33.77 & 33.78 & 33.78 \\
  \hline
   -2 & 33.76 & 33.77 & 33.77 & 33.77 & 33.78 \\
  \hline
   0 & 33.77 & 33.79 & 33.79 & 33.78 & 33.79 \\
  \hline
   2 & 33.8 & 33.83 & 33.82 & 33.81 & 33.82 \\
  \hline
   4 &  33.83 & 33.96 & 33.96 & 33.97 & 33.96 \\
  \hline
\end{tabular}

\footnotesize
\caption{WERs with different amounts of high-frequency Chinese words on `GENERAL' test set. $\theta$ is the enhancement scale in equation(\ref{eq:newweight}), $ChNum$ is the amount of high-frequency Chinese words.}
\label{tab:wer_general}
\end{table}

\begin{table}[htb]
\center
\begin{tabular}{c|c|c|c|c|c}
  \hline
    & \multicolumn{5}{c}{WER\%} \\
  \hline
   \backslashbox{$\theta$}{$ChNum$} & 1 & 2 & 3 & 4 & 5 \\
  \hline
   -4 & 66.95 & 62.68 & 60.4 & 60.68 & 61.82 \\
  \hline
   -2 & 64.39 & 63.53 & 61.54 & 61.54 & 62.68 \\
  \hline
   0 & 62.11 & 64.96 & 66.95 & 66.95 & 65.53 \\
  \hline
   2 & 62.96 & 66.95 & 65.53 & 65.53 & 65.53 \\
  \hline
   4 & 69.8 & 73.5 & 77.49 & 77.49 & 77.78 \\
  \hline
\end{tabular}

\footnotesize
\caption{WERs with different amounts of high-frequency Chinese words on `FOREIGN' test set. $\theta$ is the enhancement scale in equation(\ref{eq:newweight}), $ChNum$ is the amount of high-frequency Chinese words.}
\label{tab:wer_foreign}
\end{table}

To further examine the gains offered by the proposed approach, the name entity error rate (NEER) is used. In contrast to the WER that measures the accuracy on all words, the NEER evaluates the accuracy on focused words, i.e., the words to enhance (the new English words) and the other Chinese words in `GENERAL' test set. The results are presented in Table~\ref{tab:neer}, Table~\ref{tab:neer_ch} and Table~\ref{tab:neer_en}. It can be seen that the similar-pair-based enhancement does deliver a much better accuracy on the new English words. Importantly, the improvement on the infrequent words does not impact the performance on the other Chinese words in `GENERAL' test set, which confirms the effectiveness and safety of the proposed method. Again, the trends of NEERs on the two test sets with different values of $\theta$ and different amounts of high-frequency Chinese words are presented in Fig.~\ref{fig:wer}.

\begin{table}[htb]
\center
\begin{tabular}{l|c|c|c}
  \hline
   & &\multicolumn{2}{c}{NEER\%} \\
   \hline
   & $\theta$ & CHINESE & ENGLISH \\
   \hline
  Baseline & - & 46.85 & 100 \\
  \hline
  + SP & -4 & 48.65 & 72.09 \\
  \cline{2-4}
     & -2 & 50.45 & 48.84 \\
  \cline{2-4}
     & 0 & 50.45 & 32.56 \\
  \cline{2-4}
     & 2 & 54.05 & 0 \\
  \cline{2-4}
     & 4 & 59.46 & 0 \\
  \hline
\end{tabular}
\footnotesize
\caption{NEERs with and without the similar-pair-based enhancement. `SP' stands for enhancement with similar pairs, which use one high-frequency Chinese word. $\theta$ is the enhancement scale in equation(\ref{eq:newweight})}
\label{tab:neer}
\end{table}

\begin{table}[htb]
\center
\begin{tabular}{c|c|c|c|c|c}
  \hline
    & \multicolumn{5}{c}{NEER\%} \\
  \hline
   \backslashbox{$\theta$}{$ChNum$} & 1 & 2 & 3 & 4 & 5 \\
  \hline
   -4 & 48.65 & 49.55 & 48.65 & 48.65 & 48.65 \\
  \hline
   -2 & 50.45 & 49.55 & 51.35 & 51.35 & 51.35 \\
  \hline
   0 & 50.45 & 51.35 & 56.76 & 56.76 & 56.76 \\
  \hline
   2 & 54.05 & 58.56 & 57.66 & 57.66 & 58.56 \\
  \hline
   4 &  59.46 & 63.96 & 63.06 & 62.16 & 62.16 \\
  \hline
\end{tabular}

\footnotesize
\caption{NEERs with different amounts of high-frequency Chinese words on Chinese words of `FOREIGN' test set. $\theta$ is the enhancement scale in equation(\ref{eq:newweight}), $ChNum$ is the amount of high-frequency Chinese words.}
\label{tab:neer_ch}
\end{table}

\begin{table}[htb]
\center
\begin{tabular}{c|c|c|c|c|c}
  \hline
    & \multicolumn{5}{c}{NEER\%} \\
  \hline
   \backslashbox{$\theta$}{$ChNum$} & 1 & 2 & 3 & 4 & 5 \\
  \hline
   -4 & 72.09 & 58.13 & 53.49 & 51.16 & 48.84 \\
  \hline
   -2 & 48.84 & 30.23 & 34.88 & 32.56 & 32.56 \\
  \hline
   0 & 32.56 & 20.93 & 13.95 & 13.95 & 9.3 \\
  \hline
   2 & 0 & 0 & 0 & 0 & 0 \\
  \hline
   4 & 0 & 0 & 0 & 0 & 0 \\
  \hline
\end{tabular}

\footnotesize
\caption{NEERs with different amounts of high-frequency Chinese words on low-frequency English words in `FOREIGN' test set. $\theta$ is the enhancement scale in equation(\ref{eq:newweight}), $ChNum$ is the amount of high-frequency Chinese words.}
\label{tab:neer_en}
\end{table}

\begin{figure}[htb]
  \includegraphics[width=6cm]{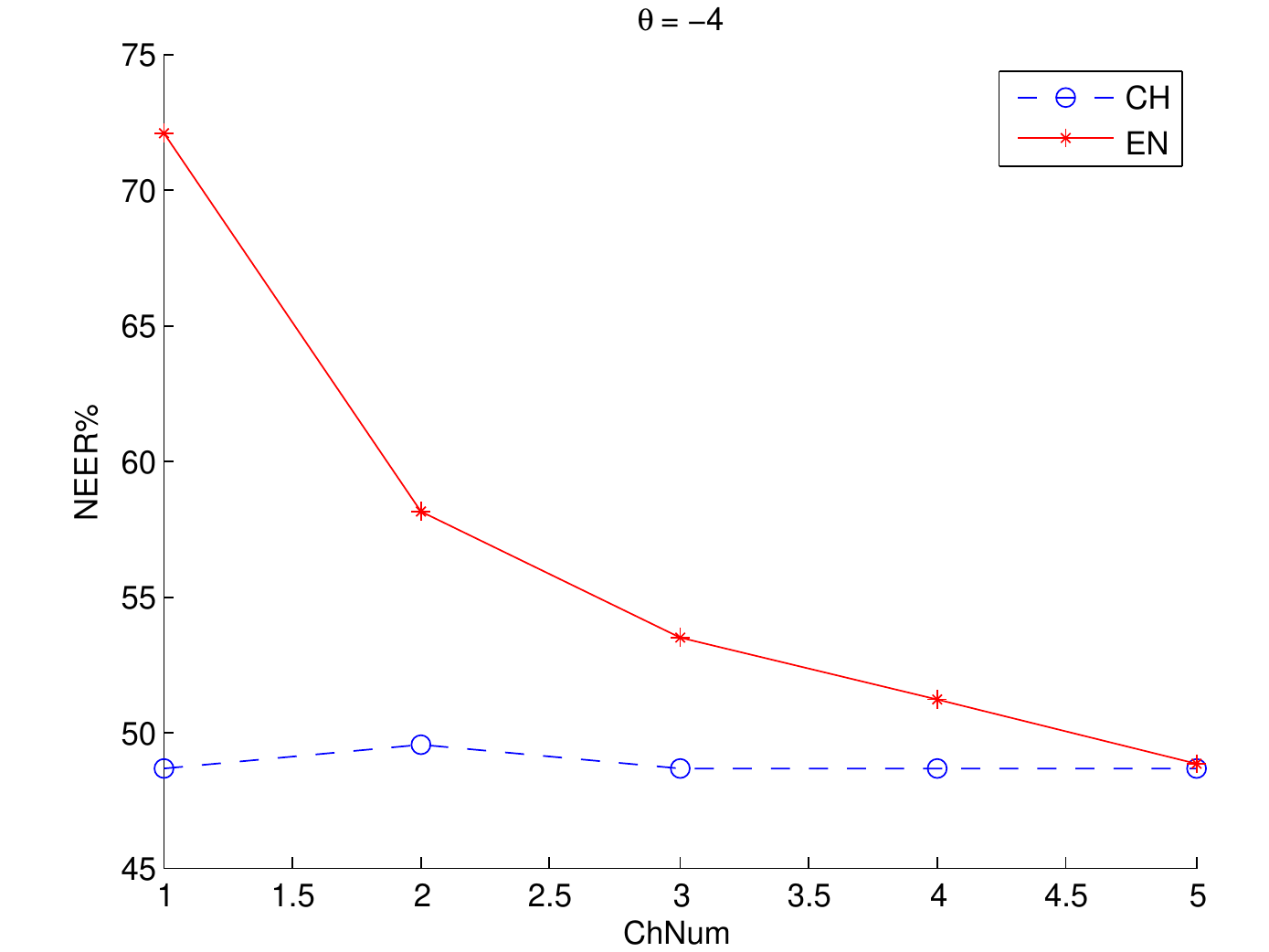}\\
  \caption{NEERs on the two types of words (low-frequency English words and other Chinese in `FOREIGN' test set) with increasing amounts of $ChNum$. The value of $\theta$ is -4}\label{fig:neer}
\end{figure}

\section{Conclusion}
\label{sec:conclusion}

In this paper, we proposed a similar-pair-based approach to enhance speech recognition accuracies on low-frequency and new words. This enhancement is obtained by exploiting the information of high-frequency words that are similar to the target words. The experimental results demonstrated that the proposed method can significantly improve performance of speech recognition on low-frequency and new words and does not impact the ASR performance in general. This lends this method to quick domain-specific adaptation where low-frequency words need to be enhanced and new words need to be supported. Future work involves enhancing low-frequency words using multiple similar words, and combining this method with other dynamic LM approaches such as the class-based LM.

\section{Acknowledgements}

This research was supported by the National Science Foundation of China (NSFC) under the project No. 61371136, and the MESTDC PhD Foundation Project No. 20130002120011. It was also supported by Sinovoice and Huilan Ltd.

\newpage
\eightpt

\bibliographystyle{IEEEtran}

\bibliography{wpair}

\begin{thebibliography}{10}
\providecommand{\url}[1]{#1}
\csname url@samestyle\endcsname
\providecommand{\newblock}{\relax}
\providecommand{\bibinfo}[2]{#2}
\providecommand{\BIBentrySTDinterwordspacing}{\spaceskip=0pt\relax}
\providecommand{\BIBentryALTinterwordstretchfactor}{4}
\providecommand{\BIBentryALTinterwordspacing}{\spaceskip=\fontdimen2\font plus
\BIBentryALTinterwordstretchfactor\fontdimen3\font minus
  \fontdimen4\font\relax}
\providecommand{\BIBforeignlanguage}[2]{{%
\expandafter\ifx\csname l@#1\endcsname\relax
\typeout{** WARNING: IEEEtran.bst: No hyphenation pattern has been}%
\typeout{** loaded for the language `#1'. Using the pattern for}%
\typeout{** the default language instead.}%
\else
\language=\csname l@#1\endcsname
\fi
#2}}
\providecommand{\BIBdecl}{\relax}
\BIBdecl

\bibitem{katz1987estimation}
S.~Katz, ``Estimation of probabilities from sparse data for the language model
  component of a speech recognizer,'' \emph{Acoustics, Speech and Signal
  Processing, IEEE Transactions on}, vol.~35, no.~3, pp. 400--401, 1987.

\bibitem{kneser1995improved}
R.~Kneser and H.~Ney, ``Improved backing-off for m-gram language modeling,'' in
  \emph{Proceedings of ICASSP}, 1995, pp. 181--184.

\bibitem{chen1996empirical}
S.~F. Chen and J.~Goodman, ``An empirical study of smoothing techniques for
  language modeling,'' \emph{Computer Speech \& Language}, vol.~13, no.~4, pp.
  359--393, 1999.

\bibitem{ward1996class}
W.~Ward and S.~Issar, ``A class based language model for speech recognition,''
  in \emph{Acoustics, Speech, and Signal Processing, 1996. ICASSP-96.
  Conference Proceedings., 1996 IEEE International Conference on},
  vol.~1.\hskip 1em plus 0.5em minus 0.4em\relax IEEE, 1996, pp. 416--418.

\bibitem{ma15}
D.~Povey, A.~Ghoshal, G.~Boulianne, L.~Burget, O.~Glembek, N.~Goel,
  M.~Hannemann, P.~Motlicek, Y.~Qian, P.~Schwarz, J.~Silovsky, G.~Stemmer, and
  K.~Vesely, ``Low-frequency word enhancement with similar pairs in speech
  recognition,'' in \emph{ChinaSIP15 (submitted)}, 2015.

\bibitem{mohri2002weighted}
M.~Mohri, F.~Pereira, and M.~Riley, ``Weighted finite-state transducers in
  speech recognition,'' \emph{Computer Speech \& Language}, vol.~16, no.~1, pp.
  69--88, 2002.

\bibitem{brown1992class}
P.~F. Brown, P.~V. Desouza, R.~L. Mercer, V.~J.~D. Pietra, and J.~C. Lai,
  ``Class-based n-gram models of natural language,'' \emph{Computational
  linguistics}, vol.~18, no.~4, pp. 467--479, 1992.

\bibitem{georges2013transducer}
M.~Georges, S.~Kanthak, and D.~Klakow, ``Transducer-based speech recognition
  with dynamic language models.'' in \emph{INTERSPEECH}.\hskip 1em plus 0.5em
  minus 0.4em\relax Citeseer, 2013, pp. 642--646.

\bibitem{schalkwyk2003speech}
J.~Schalkwyk, I.~L. Hetherington, and E.~Story, ``Speech recognition with
  dynamic grammars using finite-state transducers.'' in \emph{INTERSPEECH},
  2003.

\bibitem{dixon2012specialized}
P.~R. Dixon, C.~Hori, and H.~Kashioka, ``A specialized wfst approach for class
  models and dynamic vocabulary,'' in \emph{Thirteenth Annual Conference of the
  International Speech Communication Association}, 2012.

\bibitem{naptali2012topic}
W.~Naptali, M.~Tsuchiya, and S.~Nakagawa, ``Topic-dependent-class-based-gram
  language model,'' \emph{Audio, Speech, and Language Processing, IEEE
  Transactions on}, vol.~20, no.~5, pp. 1513--1525, 2012.

\bibitem{samuelsson1999class}
C.~Samuelsson and W.~Reichl, ``A class-based language model for
  large-vocabulary speech recognition extracted from part-of-speech
  statistics,'' in \emph{Acoustics, Speech, and Signal Processing, 1999.
  Proceedings., 1999 IEEE International Conference on}, vol.~1.\hskip 1em plus
  0.5em minus 0.4em\relax IEEE, 1999, pp. 537--540.

\bibitem{Povey_ASRU2011}
D.~Povey, A.~Ghoshal, G.~Boulianne, L.~Burget, O.~Glembek, N.~Goel,
  M.~Hannemann, P.~Motlicek, Y.~Qian, P.~Schwarz, J.~Silovsky, G.~Stemmer, and
  K.~Vesely, ``The kaldi speech recognition toolkit,'' in \emph{Proceedings of
  ASRU}, 2011, pp. 1--4.

\end{thebibliography}

\end{document}